\pdfoutput=1
\documentclass[11pt]{article}

\usepackage{acl}

\usepackage{times}
\usepackage{latexsym}

\usepackage[T1]{fontenc}
\usepackage[utf8]{inputenc}

\usepackage{microtype}
\usepackage{svg}

\usepackage{graphicx}
\usepackage{booktabs}
\title{Assessing Group-level Gender Bias in Professional Evaluations: The Case of Medical Student End-of-Shift Feedback}

\author{Emmy Liu \\
  Language Technologies Institute \\
  Carnegie Mellon University \\
  \texttt{mengyan3@cs.cmu.edu} \\\And
  Michael Henry Tessler \\
  MIT \\
  Brain and Cognitive Sciences \\
  \texttt{tessler@mit.edu} \\\AND
  Nicole Dubosh \\ 
  Beth Israel Deaconess Medical Center \\
  Harvard Medical School \\
  \texttt{ndubosh@bidmc.harvard.edu} \\ \And
  Katherine Mosher Hiller \\
  Indiana University \\
  School of Medicine, Bloomington \\
  \texttt{kmhiller@iu.edu} \\
  \AND
  Roger P. Levy \\
  MIT \\
  Brain and Cognitive Sciences \\
  \texttt{rplevy@mit.edu}
  }

\begin{document}
\maketitle
\begin{abstract}
 Although approximately 50\% of medical school graduates today are women, female physicians tend to be underrepresented in senior positions, make less money than their male counterparts and receive fewer promotions. There is a growing body of literature demonstrating gender bias in various forms of evaluation in medicine, but this work was mainly conducted by looking for specific words using fixed dictionaries such as LIWC and focused on recommendation letters. We use a dataset of written and quantitative assessments of medical student performance on individual shifts of work, collected across multiple institutions, to investigate the extent to which gender bias exists in a day-to-day context for medical students. We investigate differences in the narrative comments given to male and female students by both male or female faculty assessors, using a fine-tuned BERT model. This allows us to examine whether groups are written about in systematically different ways, without relying on hand-crafted wordlists or topic models. We compare these results to results from the traditional LIWC method and find that, although we find no evidence of group-level gender bias in this dataset, terms related to family and children are used more in feedback given to women. 
\end{abstract}

\section{Introduction}
Female physicians and trainees have advanced considerably in the medical field within recent years, and approximately 50\% of medical school graduates are now women \cite{women-in-medicine-1}. However, female physicians lag their male counterparts in salary, promotions, and positions in senior leadership \cite{women-in-medicine-1, women-in-medicine-2, women-in-medicine-3, women-in-medicine-4}. A mechanism that perpetuates this inequality may be unequal evaluations of male and female physicians. Past work has revealed gender bias in several forms of evaluation. Evaluations of recommendation letters in academia found that women tended to be described in communal traits (caring, nurturing) whereas men were described in agentic terms (ambitious and self-confident) \cite{rec-letters}. The same trend holds in direct observation comments given to Emergency Medicine (EM) residents, with feedback themes varying by gender, particularly around the domains of authority and assertiveness \cite{em-evals}. In the same context, women were also found to receive more contradictory and polarized assessments on their skills as compared to men \cite{em-evals}. 

If there are systemic differences in evaluations for different genders, it may be possible that these differences arise early in a student's career and snowball into fewer opportunities in late career, when they are quantitatively detectable through metrics such as salary and number of promotions. It is important to understand at what phases of a student's career inequities arise, so that interventions can be targeted toward supporting women or other underrepresented minorities at these stages. Focus groups of female physicians in the field find different experiences at early, mid, and late career stages, with older women experiencing more overt discrimination, and younger women reporting more implicit bias, though it is unknown if this is due to decreased discrimination in recent years, or due to younger physicians not yet recognizing signs of discrimination \cite{women-physician-careers}.  

Findings on gender differences in language are mixed for students earlier in their careers. A qualitative analysis of surgical residency letters of recommendation, collected from before the students applied for residency, found that male applicants' letters contained more achievement-oriented terms, whereas female applicants' letters contained more care-oriented terms \cite{early-evals-1}. However, a similar analysis on the EM standardized letter of evaluation found no such difference \cite{early-evals-2}. 

To investigate this question more thoroughly, we use a new dataset of written assessments on medical students' work based on individual shift performance before their residencies. Most previous work from the medical community has used relatively simple linguistic methods such as the Linguistic Inquiry and Word Count dictionary (LIWC) \cite{liwc, rec-letters, early-evals-2, early-evals-3}, but using pretrained language models may allow us to investigate bias in a more fine-grained manner \cite{other-nlp, other-nlp-2}. Additionally, existing work on medical bias within the NLP community mainly focuses on patients, rather than physicians themselves \cite{nlp-clinical}.

We fine-tune a pretrained BERT model and use its predictions as a tool to try to identify group-level prediction residuals. If such a difference exists on a systematic level, it may indicate that assessors are writing about students in different ways based on their gender, given the same objective performance. Caution should be taken when using similar methods as language models can also come imbued with biases of their own, but we outline the method in this work and highlight its use in comparing model predictions and human judgments when both text data and quantitative data are available. 

Although we can replicate past work showing a significant difference in social-communal terms used to describe women, we do not find as clear a relationship between comments written about a student and the global score given on a shift. We do not find a systemic difference between male and female students when comparing group-wide residual differences. This indicates that although male and female students may be written about differently, no gender is written about in a systemically worse way. Due to privacy concerns, the dataset is not available online, but the full dataset can be obtained through emailing our medical co-author: \texttt{kmhiller@iu.edu}. 

\section{Bias Statement}

We study the relationship between text comments and numeric ratings of performance given to male and female medical students. We introduce the method of comparing language model residual predictions to numeric data to find group-wide differences in language use. We fine-tune a language model to predict the rating associated with a given comment about a student, and ask if there is a cross-group difference in the residual error that the trained model makes. For instance, are female students given less positive-sounding comments than their male counterparts for the same level of clinical skills (as measured by their numeric evaluation scores)?

Feedback from supervisors is used to make decisions on whether a student receives a residency, or later on whether they get promoted to a higher position within medicine. This has potential to address allocational harms to women within medicine. The under-representation of women in senior positions in medicine could also lead to wider harms in inequity as a result.

There are shortcomings in presenting gender as a binary, and in this dataset gender information was not collected based on self-identification. We hope that future work will explore a wider diversity of gender identification, but we present this analysis as a first step.

\section{Dataset}

The dataset consists of evaluations of undergraduate medical students conducted with the National Clinical Assessment Tool (NCAT-EM), the first standardized assessment based on direct observation of skills in a clinical setting \cite{ncat-1, ncat-2}. The NCAT-EM was developed by EM educators, and has been implemented at 13 institutions in the United States. Data was collected from departments participating in the NCAT-EM Consortium from 2017-2019 \cite{ncat-2}.

\begin{figure}[!ht]
    \centering
    \includegraphics[scale=0.25]{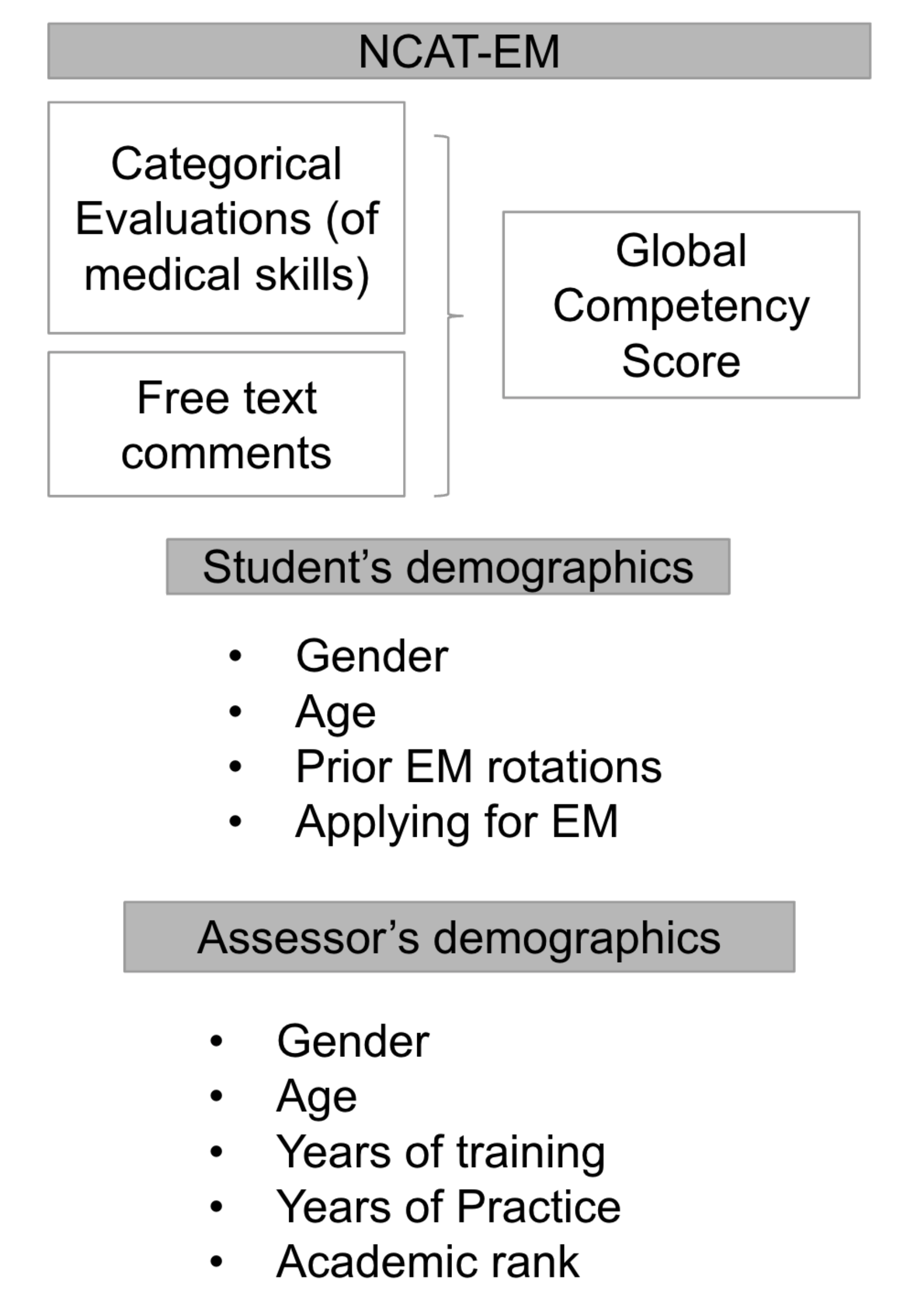}
    \caption{Data and features included in the NCAT-EM dataset.}
    \label{fig:data-features}
\end{figure}

The dataset contains short free text comments on a student's performance, categorical assessments on multiple skill areas, a global competency score (lower third, middle third, top third, and top 10\%), as well as demographic information about students and assessors: gender, age, rank of assessor (junior vs senior faculty). These attributes are outlined in \autoref{fig:data-features}. Examples of free-text comments and associated scores are given in \autoref{tab:comment-examples}.

\begin{table}[!ht]
    \small
    \begin{tabular}{p{0.22\linewidth}p{0.66\linewidth}}
      \toprule
      Global Score &  Comment \\\midrule
       0 & always seemed happy to help but did not reassess patients or follow up labs on own. also, came off as arrogant to multiple residents in the department. \\
      \midrule
      1 & good differential, interested, team player.  \\
      \midrule
       2 & great job keeping up with your patients; we had a very sick one and you made sure you were on top of it. \\
      \midrule
       3 & i really enjoyed working with x, as x was a very thorough historian, and provided a brief focused history. x appears to have a good grip of emergency medicine at this point, and provides a good reasonable plan. x is going to be an exceptional resident, and will continue to improve significantly over the next year. \\
      \bottomrule
    \end{tabular}
     \caption{Examples of free text comments written about students (after preprocessing), with the associated global competency score (on a scale from 0 to 3).}
    \label{tab:comment-examples}
  \end{table}

After excluding samples with missing data, there were 3162 individual assessments, where 1767 were evaluations of male students and 1395 were evaluations of female students. Because students may work multiple shifts, and the same supervisor may supervise multiple students, there are some students and assessors who are repeated, although each sample represents a different shift. Names and named entities were removed from comments using the spaCy entity recognizer and replaced with the letter "x". Gendered pronouns were removed and replaced with the gender-neutral pronoun "they".

This dataset consists mostly of short comments focused on student performance. The mean number of words in a comment was 28.4, and the maximum number of words in a comment was 187. 
%The overall distribution of assessment ratings is illustrated in \autoref{fig:score-dist}. 
The overall distribution of assessment ratings was: 5\% in bottom third, 35\% in middle third, 45\% in top third, and 15\% in the top 10\%. A slightly higher density of female students received the top rating compared to male students. We convert these to integer values from 0-3. 

%\begin{figure}
    %\centering
    %\includegraphics[scale=0.6]{score-distribution.eps}
    %\caption{Distribution of global competency scores given to male and female students.}
    %\label{fig:score-dist}
%\end{figure}
%\section{Methods}

We use two main methods to identify possible biases in this dataset: prediction residual analysis and word/topic based analysis. Previous work has focused on word-level analysis, but since we have access to both comments as well as a competency score, we investigate to what extent we can reconstruct the mapping from text comment to the score a student receives by applying a language model, and if there are differences in this mapping between male and female students. We used 70\% of the dataset to train, 15\% to evaluate, and 15\% as the test set.

\subsection{Language Model Prediction Residuals}

\begin{figure*}[!ht]
    \centering
    \includegraphics[width=0.8\textwidth]{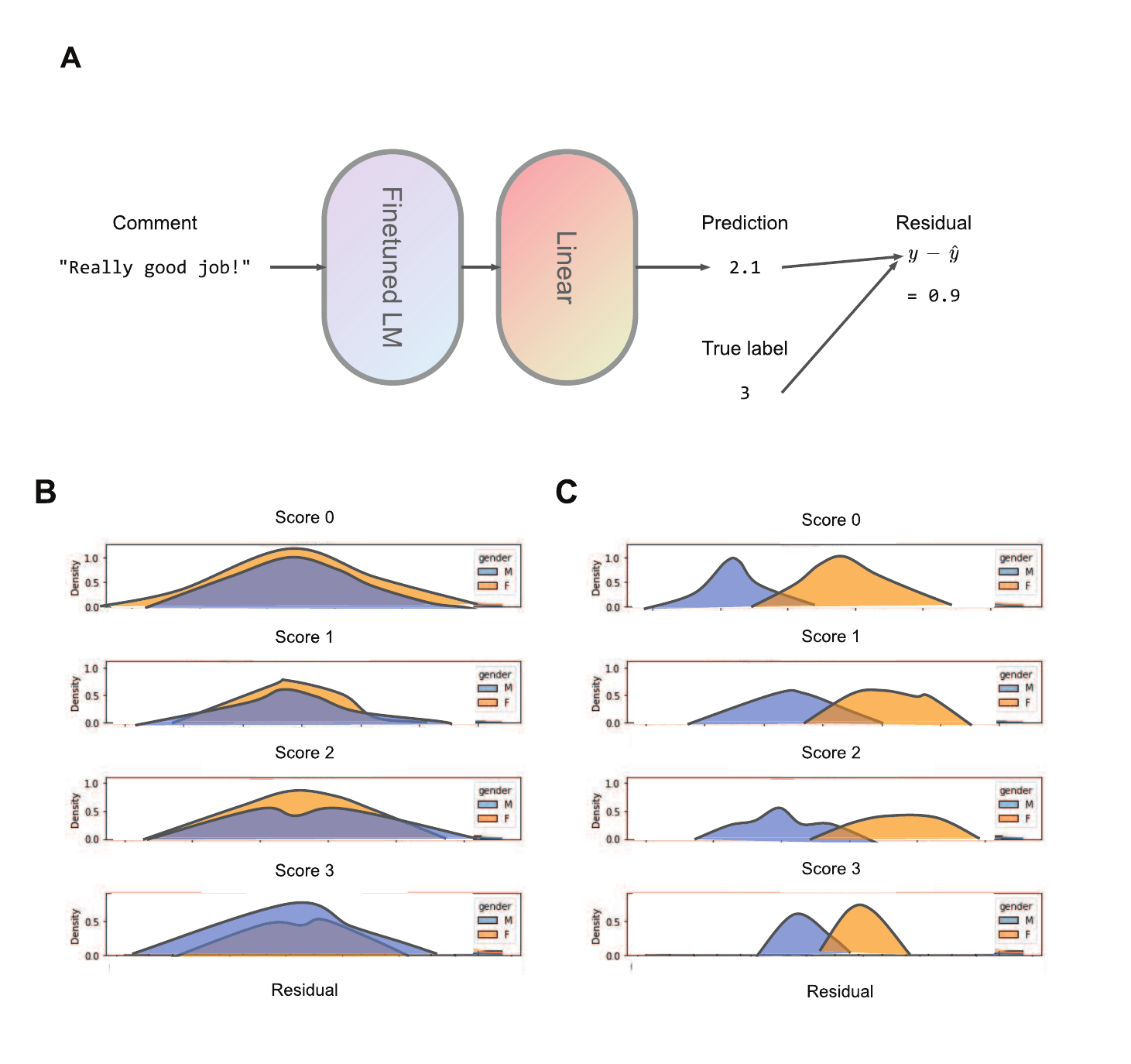}
    \caption{Panel A illustrates the LM residual method (for illustration purposes, the areas under the curve in this drawing are not necessarily the same as they would be in reality). A language model is finetuned on text evaluations without gender information to predict the global rating. Panel B illustrates a case with no differences in residuals between the male and female group, illustrating the case without textual bias. Panel C illustrates a biased case. In this hypothetical case, female students received a score that is consistently higher than the language in their comments would suggest.}
    \label{fig:setup}
\end{figure*}

In order to examine the relationship between text comments and the global competency score, we finetune  \texttt{bert-base-uncased} with early stopping on the free text comments with gender and institution information removed, with a linear layer trained to predict the global competency score \footnote{We used the Adam optimizer with a learning rate of \texttt{5e-6}, epsilon of \texttt{1e-8}, and weight decay \texttt{1e-10}, and a batch size of 32. These parameters were taken from the default settings of the transformers implementation of Adam at the time, with a minimal hyperparameter search over learning rate.}. We then examine the prediction residuals of the finetuned model on a group level for group $\mathcal{C}$: 

\begin{equation}
    \delta_{\mathcal{C}} = \{ y_i - \hat{y_i}\}_{i = 0}^{|\mathcal{C}|}
\end{equation}

In the student-only setup the groups would be the set of male students, $\mathcal{M}$, and the set of female students $\mathcal{F}$. In the student and assessor setup, the groups would be the different assessor and student gender combinations, namely $\mathcal{M} \times \mathcal{M}$, $\mathcal{M} \times \mathcal{F}$, $\mathcal{F} \times \mathcal{M}$, and $\mathcal{F} \times \mathcal{F}$. The null hypothesis is that there should be no difference between these groups, for instance $\delta_{\mathcal{F}}$ and $\delta_{\mathcal{M}}$. If there is a significant difference, it indicates that there may be a difference in the relationship between text and global score between these groups. For instance, if the $\delta_{F_i}$s are significantly higher, this would indicate that scores given to female students are significantly higher than expected given comments about them. Note that $\hat{y_i}$ is based on text from which explicit gender markers were removed.

\subsection{LIWC}
In order to check if previous results using LIWC replicated on this dataset, we examined many categories of words from LIWC \footnote{Specifically, we examined these categories: Affect, Positive Emotion, Negative Emotion, Social, Cognitive Processing, Insight, Achieve, Standout, Ability, Grindstone, Teaching, Research, Communal, Social-Communal, and Agentic. The associated words can be found in either the standard LIWC dictionary or in these references: \cite{liwc, rec-letters}}. 

Additionally, we used user-defined dictionaries from previous studies of letters of recommendation: grindstone words (e.g. diligent, careful), ability words (e.g. talented, intelligent), standout adjectives (e.g. exceptional), research terms (e.g. research, data), teaching terms (e.g. teach, communicate), social-communal terms, (e.g. families, kids), and agentic terms (e.g. assertive, aggressive) \cite{rec-letters, liwc-1, liwc-2, liwc-3, liwc-4}. The prevalence of these categories was found to differ in past studies of recommendation letters. We used a coding scheme of 0 if a theme did not show up in a comment, and 1 if it did. We used a Fisher exact test on comments written about male or female students, with Holm-Bonferroni correction to control for multiple comparisons.

\section{Results}

\subsection{Residual Analysis}
We present the results for the residual analysis first. We note first that the language model achieved relatively low accuracy when its predictions were rounded to the nearest integer (46\%), but made comparable predictions to humans (50.7\% average across three annotators, on a randomly-sampled 20\% of the dataset. The annotator agreement was moderate (Krippendorff's $\alpha$ = 0.491). \footnote{Annotators were shown a text comment and assigned a global rating from 0-3. They did not see the labels for that portion of the dataset, but were allowed to look at labels for the remaining 80\% to guide their judgment. Additionally, annotators were all familiar with the dataset and rubric for global score.}), This indicates a noisy mapping between text and global score in this dataset.

\begin{figure}[!ht]
    \centering
    \includegraphics[width=\columnwidth]{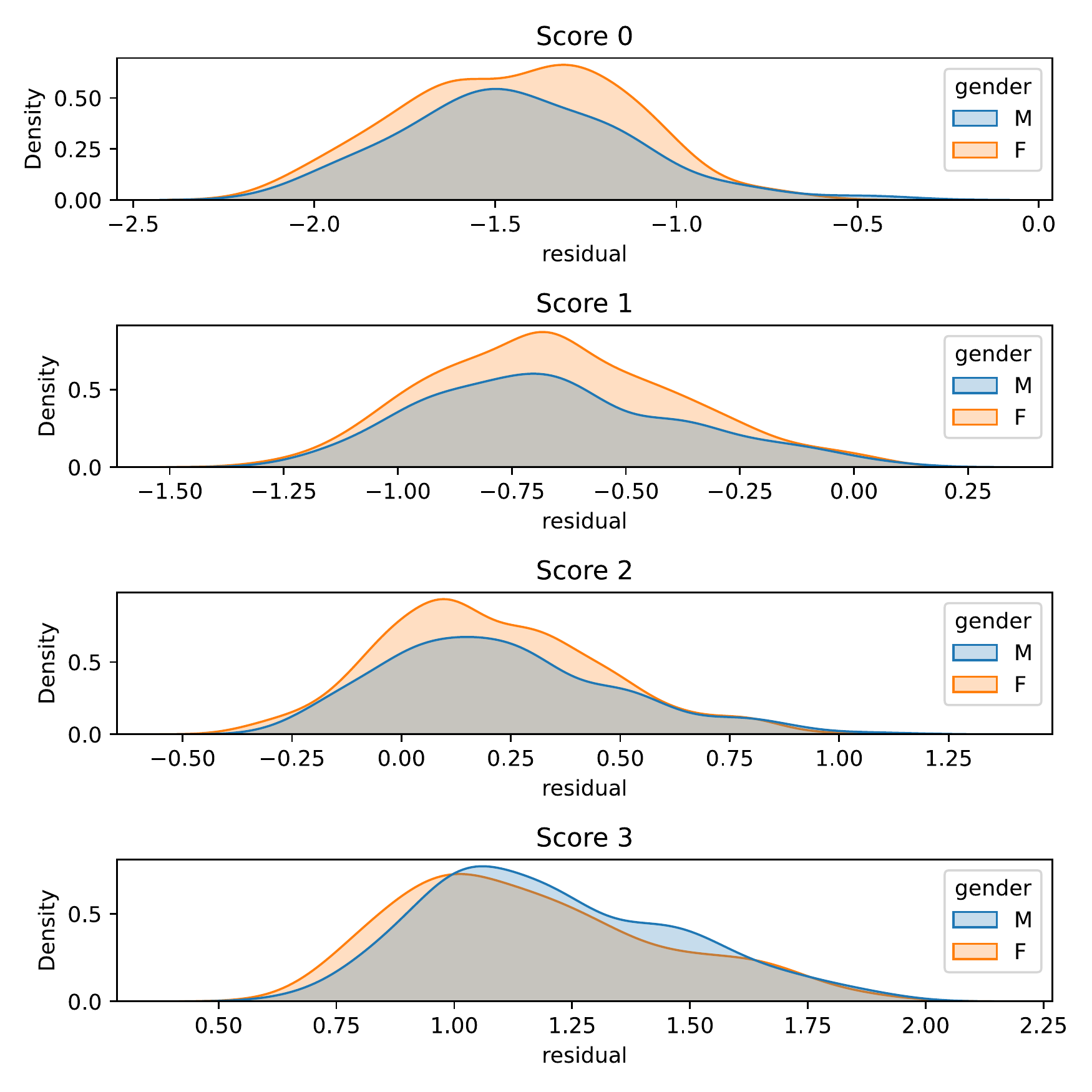}
    \caption{Residual densities for male and female students by global score (0-3). BERT did not achieve a high accuracy on this task, but there was no significant difference in group-wise residuals, showing that male and female students tended to receive comments of a similar valence for their associated score.}
    \label{fig:results}
\end{figure}

Results following the format of \autoref{fig:setup} are found in \autoref{fig:results}, and visual inspection does not reveal differences in residuals. A T-test comparing male and female global scores in the entire dataset confirmed that female students had a slightly higher score (higher mean by 0.08, $p < 0.004$). However, no significant difference was found between residuals for male and female students ($p = 0.517$). There were also no significant differences between BERT predictions themselves for male and female students ($p = 0.152$). 

When considering both assessor gender and student gender, we performed an ANOVA test and found that two groups had statistically significant differences in means: when comparing male assessor, male student pairs with male assessor, female student pairs, the male assessor, female student pairs had marginally higher ratings (0.0893 difference, $p = 0.0485)$). When comparing male assessor, male student pairs with female assessor, female student pairs, female assessor and female student pairs also had higher mean ratings (0.114 difference, $p = 0.0411$). These effects are more marginal, but expected given the slightly higher scores of female students. 

When examining the residuals for the 4-way split, there was one statistically significant difference, between male assessor, female student pairs and female assessor, female student pairs. The residual mean was 0.1195 higher in male assessor, female student pairs, and this was significant to a marginal degree ($p = 0.0468$). This indicates that the actual score given by male assessors to female students was higher than their comment would suggest, as compared to female assessors giving comments to female students. However, this was a marginal effect, and overall we find no clear evidence of gender bias in the comments given to students, or the relationship of the comments given to global score received. 

\subsection{LIWC}

We examine LIWC themes by student gender and partially replicate previous results showing that women tend to be described with more social-communal language than men \cite{rec-letters}. We did not find any significant differences when dividing by assessor gender. However, we did not find that women were described as less agentic in this dataset. A summary of the percentage of comments in which themes occurred is summarized in \autoref{tab:themes}. Only the difference in the Social-communal theme is highly significant ($p < 6 \times 10^{-16}$) after Holm-Bonferroni correction. This theme consists of family terms (families, babies, kids), e.g. "great proactive attitude in approaching members of the team and interacting with patients and their families". There is some variation in these comments, as some concern bedside manners with patients and families, and some comment on ability to work with children, which may be necessary in a pediatric unit. We did not see a significant difference in the communal theme (which would describe a warm and nurturing student), unlike in past work. 

\begin{table*}[!ht]
    \centering
    \small
    \begin{tabular}{*{6}{p{.15\linewidth}}}
    \toprule
        Theme & Example words & \% comments with theme (M) & \% comments with theme (F) & odds ratio (M/F) & $p$ (corrected) \\
    \midrule
        Affect & amazing, arrogan*, apath*, interest & 93.1\% & 92.7\% & 1.11 &  1\\
    \midrule
        Positive Emotion & fantastic, improve, brilliant & 89.5\% & 89.4\% & 1.03 & 1 \\
    \midrule
        Negative Emotion & angry, difficulty, fail & 42.67\% & 41.36\% & 1.06 & 1\\
    \midrule
        Social & advice, ask, commun* & 59.9\% & 57.9\% & 1.08 & 1\\
    \midrule
        Cognitive Processing & accura*, inquir*, interpret* & 76.6\% & 75.0\% & 1.09 & 1\\
    \midrule
        Insight & deduc*, explain, reflect* & 53.6\% & 52.2\% & 1.05 & 1\\
    \midrule
        Achieve & abilit*, ambition, leader* &  67.1\% & 66.7\% & 1.02 & 1\\
    \midrule
        Standout & outstanding, exceptional, amazing & 17.0\% & 20.6\% & 0.786 & 0.1309 \\
    \midrule
        Ability & talen*, smart, skill & 18.4\% & 19.1\% & 0.960 & 1\\
    \midrule
        Grindstone & reliab*, hardworking, thorough & 45.3\% & 46.2\% & 0.965 & 1\\
    \midrule 
        Teaching & teach, mentor, communicate* & 21.1\% & 22.7\% & 0.914 & 1\\
    \midrule
        Research & research*, data, study & 9.96\% & 9.75\% & 1.02 & 1\\
    \midrule 
        Communal & kind, agreeable, caring & 4.07\% & 4.87\% & 0.829 & 1\\
    \midrule
        Social-communal & families, babies, kids & 8.26\% & 18.4\% & 0.401 & $5.88 \times 10^{-16}$* \\
    \midrule 
        Agentic (adjectives) & assertive, confident, dominant & 1.75\% & 1.72\% & 1.02 &  1\\
    \midrule 
        Agentic (Orientation) & do, know, think & 10.2\% & 7.67\% & 1.37 & 0.1789 \\
    \bottomrule

    \end{tabular}
    \caption{LIWC theme occurrence in comments given to male and female students. A higher percentage of comments contained the social-communal theme for women than for men. $p$-values were corrected with the Holm-Bonferroni correction.}
    \label{tab:themes}
\end{table*}
\section{Conclusion}

Gender bias in medical education is a major barrier to women in the field, and it is important to know in what circumstances and career stages it occurs in order to create targeted training and intervention. Previous work has found that there may be potential bias in medical student recommendation letters, but we investigate whether there is systemic bias in an everyday setting in feedback given to male and female medical students. We collect data using NCAT-EM evaluations to answer this question, and use language model residuals to investigate the relationship between free text comments and integer ratings given to students. We find no evidence of bias using the residual definition, although we find that there is a statistically significant difference in the percentage of comments that mention social-communal themes, with women receiving more mentions of family-oriented words in their evaluations. 

One limitation of this dataset is that the mapping between text comment and global score is quite noisy, as neither a fine-tuned language model nor human judges were able to achieve a high score in classifying the text based on the global rating. However, the prediction residual method can be used in any dataset with both text data and outcome data, for instance applications to educational programs, or employee evaluations. One caveat is that language models themselves can be biased, so this method is best applied after sensitive attributes have been obfuscated.

Additionally, this dataset is quite small and limited to a relatively small set of samples. It is possible that biases could be found in a larger dataset of shift evaluations, or in data collected from a different set of institutions. However, we leave such data collection to future work, and hope that this encourages the collection and analysis of similar data on a wide scale. We hope that this work will inspire further research into how bias manifests or does not manifest at different stages of professionals' careers, and how we can combine multiple sources of information together with text to form a wider view of bias and fairness.

%Note that this method also quantifies an interaction between human and language model bias, and can be used post-hoc to quantify language model bias as well. For instance, if a resume filtering system is suspected of being biased, a trained human team can label the quality of the anonymized applications, and the group-wise prediction residual between the humans making predictions on anonymized data and the model making predictions on de-anonymized data can be compared. 

\section*{Acknowledgements}
We thank Lynnette Ng and Samantha Phillips for participating in the human annotation task. % Funding information here?

\bibliography{anthology,custom}
\bibliographystyle{acl_natbib}

\end{document}